\documentclass{article}

\usepackage{arxiv}

\usepackage[utf8]{inputenc} % allow utf-8 input
\usepackage[T1]{fontenc}    % use 8-bit T1 fonts
\usepackage{hyperref}       % hyperlinks
\usepackage{url}            % simple URL typesetting
\usepackage{booktabs}       % professional-quality tables
\usepackage{amsfonts}       % blackboard math symbols
\usepackage{nicefrac}       % compact symbols for 1/2, etc.
\usepackage{microtype}      % microtypography
\usepackage{lipsum}		% Can be removed after putting your text content
\usepackage{graphicx}
\usepackage{bm}
\usepackage{amsmath, amssymb}
\newcommand{\argmax}{\mathop{\rm arg~max}\limits}

\title{Selecting Data Adaptive Learner from Multiple Deep Learners using Bayesian Networks}

\author{ {\hspace{1mm}Shusuke Kobayashi} \\
	Department of Medicine\\
	Chiba University\\
	1-8-1 Inohana, Chuo-Ku, Chiba 260-8670, Japan \\
	\texttt{shusuke.kbys@chiba-u.jp} \\
	\And
	{\hspace{1mm}Susumu Shirayama} \\
	Graduate School of Engineering\\
	The University of Tokyo\\
	7-3-1 Hongo, Bunkyo-Ku, Tokyo 113-8656, Japan\\
	\texttt{shirayama.susumu@gmail.com} \\
}

\date{}

\renewcommand{\headeright}{Preprint Submitted to Neural Computing and Applications}
\renewcommand{\undertitle}{Preprint Submitted to Neural Computing and Applications}

\hypersetup{
pdftitle={Selecting Data Adaptive Learner from Multiple Deep Learners using Bayesian Networks},
pdfsubject={cs},
pdfauthor={Shusuke Kobayashi, Susumu Shirayama},
pdfkeywords={Time series forecasting, Deep Learning, Bayesian Network, Mixture of Experts},
}

\begin{document}
\maketitle

\begin{abstract}
A method to predict time-series using multiple deep learners and a Bayesian network is proposed. In this study, the input explanatory variables are Bayesian network nodes that are associated with learners. Training data are divided using K-means clustering, and multiple deep learners are trained depending on the cluster. A Bayesian network is used to determine which deep learner is in charge of predicting a time-series. We determine a threshold value and select learners with a posterior probability equal to or greater than the threshold value, which could facilitate more robust prediction. The proposed method is applied to financial time-series data, and the predicted results for the Nikkei 225 index are demonstrated.
\end{abstract}

% keywords can be removed
\keywords{Time series forecasting \and Deep Learning \and Bayesian Network \and Mixture of Experts}

\section{Introduction}
In the course of developing deep learning factors such as the acquisition and generation of appropriate training data, the long calculation time required for learning, and difficulties selecting parameters are often highlighted.Research to solve such problems is ongoing \cite{Bengio2012},\cite{Hutter2016},\cite{Lorenzo2017},\cite{Snoek2012},\cite{Kuremoto2014}, although implications of the results of such research has not been explained. However, in recent years, many reports have suggested that substitution into deep learning is progressing. Such reports focus on the application of neural networks to improve efficiency and accuracy \cite{Bengio2007},\cite{Krizhevsky2012},\cite{Dahl2012}.

Methods that involve multiple learners, such as ensemble and complementary learning, greatly contribute to improvements in efficiency and accuracy. In ensemble learning, outputs from learners are integrated by weighted average or voting methods \cite{Wang2017},\cite{Suk2017},\cite{Zhao2017}. Complementary learning combines learners to compensate for each learner's disadvantages \cite{nomiya2007}. The learners used in these methods are primarily weak learners. Note that many tasks can be divided into subtasks; thus, a hierarchical control mechanism can be adopted. To realize a hierarchical control mechanism, multiple learners can be used in some cases.

Takahashi and Asada proposed a robot behavior acquisition method by hierarchically constructing multiple learners of the same structure \cite{takahashi2000}. Herein, a learner that allows lower level learners to take charge of different subtasks and learn low level actions makes upper level learners learn higher level actions using lower level learning devices.

Although differing from ensemble learning and complementary learning, the Mixture of Experts (MoE) technique \cite{jacob1991} is a prediction model that uses multiple learners. MoE is based on the divide-and-conquer principle wherein prediction is performed by dividing the input data space into several small areas and assigning a single neural network to each divided area. Rather than learning one large neural network, the MoE model attempts to increase learning efficiency by dividing complex problems into smaller areas.

MoE combines the outputs of several expert networks, each of which focuses on a different part of the input space. This is achieved through a learned gating network that integrates the outputs of expert networks to produce a final output. Different types of expert architectures have been proposed, such as SVMs \cite{collobert2002}, Gaussian processes \cite{tresp2000}\cite{theis2015}\cite{deisenroth2015}, Dirichlet processes \cite{shahbaba2009}, and deep learners. Since its introduction more than 20 years ago, MoE has been studied extensively, and various implementation have been proposed \cite{eigen2014}\cite{shazeer2017}\cite{sam2017}.

We proposed a learning framework that combines multiple deep learners and naive Bayes' classifier to select the appropriate learner \cite{JDAIP2017}.  In this paper, we propose a more accurate prediction method using Bayesian networks.
The proposed method has the following different features compared to \cite{JDAIP2017}.
First, It is possible to visualize the relationship between explanatory variables and predict future value at the same time. Next, since a Bayesian network is used, when a deficiency occurs in an explanatory variable, it is possible to estimate a deficient explanatory variable from other variables. This property is important for prediction in areas such as medicine and economics where complete data is not always available. The framework proposed in this paper is general and robust to data, so it can be applied to various other AI fields.

The proposed method divides training data using K-means clustering and creates multiple deep learners using the divided training data. We used a Bayesian network to select a learner suitable for prediction. Although K-means is a deterministic method, the Bayesian network can select the learner in consideration of probability and visualize relationships between variables.We also consider incorporating the probability output from the Bayes classifier into the prediction. We determine a threshold value and select multiple learners with a posterior probability equal to or greater than the threshold value, which facilitates more robust prediction. 

The proposed method is similar to the ensemble learning bagging method such that the training data are divided and each division is learned independently. The differences between the proposed method and ensemble learning include the division method and the method used to integrate multiple learners, i.e., the learner selection method.

Compared to Takahashi and Asada's ideas, tasks do not need to be divided into subtasks; the learner selection method is different. In addition, Takahashi and Asada used Q-learning extended to continuous state behavior space as a learning device \cite{takahashi1999} whereas we used deep learning.

The proposed method is also similar to MoE in that it divides data and creates a learner for each cluster. However, it differs in that a Bayesian classifier, in particular a Bayesian network, is used to select learners. To the best of our knowledge, no study has employed a Bayesian classifier for a gating network. In the original MoE model \cite{jacob1991}, the number of expert networks is determined in advance, and the expert and gating network parameters are determined sequentially using an EM algorithm. With the proposed method, we determined the number of expert networks in advance as well as dynamically according to the data. Furthermore, a Bayesian classifier that assumes the role of a gating network is learned from the K-means results, and expert networks are trained for each K-means cluster.

The proposed method simply creates multiple learners using the divided training data. In addition, we used a simple K-means method to divide the data, and we cannot examine the differences in the result via other division methods. We used a Bayesian network to integrate multiple deep learners and selected deep learners suitable for prediction; however, because we used the simplest network, we cannot solve the structural complexity of the Bayesian network itself. In the Bayesian network, we express random variables as nodes and quantitative dependencies between variables as conditional probabilities. In this case, determining the structure of the network in advance is required; however, it may be possible to adopt various network structures. 

In this study, we begin by presenting a more general framework to learn multiple deep learners and select appropriate learners suitable for prediction using Bayesian networks. Specifically, we assume that multiple deep learners are generated from different types of data and that the network structure of the Bayesian network is determined such that it is compatible with the data. Then, the Bayesian network is used to select the learner suitable for prediction. In this study, the Nikkei Average Stock Price Forecast, which considers the influence of multiple stock markets, is taken as a case. Specifically, we estimate the Nikkei Stock Average for this term from the Nikkei Stock Average of the previous term and major foreign stock price indicators, such as the NY Dow Jones Index and the FTSE 100. We evaluate the validity of the proposed method based on the accuracy of the estimation result.

\section{Approach}
\subsection{Concept}
This section discusses the concept behind the proposed method. 
Training data $\bm{S} \in \mathbb{R}^{n \times d}$ are divided into $K$ clusters. Learning $K$ deep learners is performed independently with each dataset. We consider the problem of obtaining predictions $\hat{\bm{y}} \in \mathbb{R}^{m}$ from test data $\bm{S}_{test} \in \mathbb{R}^{m \times d}$ with $K$ deep learners $l_{1},...,l_{K}$.

A Bayesian network can be combined with multiple deep learners using two approaches. One approach involves applying a Bayesian network prior to inputting data to multiple deep learners, and the other involves applying a Bayesian network when integrating the outputs from multiple learners. In the first approach, the explanatory variables of input data are the nodes of a Bayesian network and are associated with learners. In the second approach, the outputs of all learners are considered the nodes of the Bayesian network and are integrated. We employ the former approach.

In the proposed method, a Bayesian network is applied to inputs $\bm{x} = (x_1,...,x_d)^{\mathrm{T}}$, and the probability is obtained. The obtained probability indicates which learner is appropriate for association with input $\bm{x}$. If the dimension of the input data is $d$, the Bayesian network has $d$ nodes that indicate each variable $x_{1}, ..., x_{d}$ of the input data. Here one node represents the selected learner, and the total number of nodes is $d + 1$. The Bayesian network is learned from the training data and clustering result. The prediction $\hat{y} \in \mathbb{R}$ is obtained from the output of all learners $l_{1},...,l_{K}$ and the probability from the Bayesian network.

\begin{figure*}[!h]
  \begin{center}
    \includegraphics[clip,width=11.0cm]{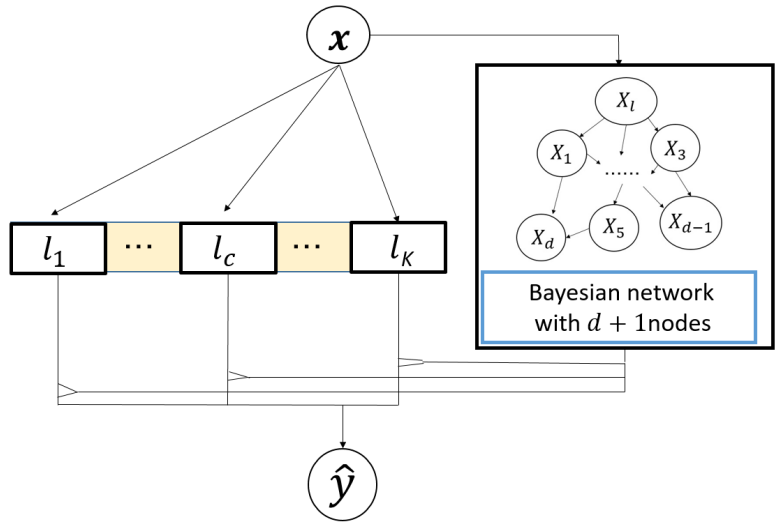}
    \caption{Outline of the proposed method}
    \label{fig:methodA}
  \end{center}
\end{figure*}

Figure\ref{fig:methodA} shows an outline of the proposed method. Here, $X_{l}$ is the random variable representing the selection of a learner that is suitable for prediction. The process of the proposed method is summarized as follows.

\begin{enumerate}
\item Let $\bm{S} \in \mathbb{R}^{n \times d}$ be the training data and $\bm{y} \in \mathbb{R}^n$ be the label data. 

Training data $\bm{S}$ are divided into $K$ clusters. 
The division number $K$ is determined dynamically according to the data\cite{pelleg}. 
The training data for each cluster, the corresponding label data $(\bm{S}_c,\bm{y}_c )_{c = 1}^{K}$, and vector $\bm{c}$ (stores the cluster index to which each row of $\bm{S}$ belongs) are obtained. $l_c \; (c = 1,..,K)$ are trained with $(\bm{S}_c,\bm{y}_c )$.

\item 
The Bayesian network is trained with $(\bm{S},\bm{c})$. 
First, the structure of the Bayesian network is determined heuristically from the data based on the information criterion. In this paper, six Bayesian networks are constructed using six different methods. Subsequently, the conditional probability can be estimated.

\item 
Let $\bm{S}_{test} \in \mathbb{R}^{m \times d}$ be test data.\\
$\bm{S}_{test}$ are the inputs to the Bayesian network, and the posterior probability $\bm{P} \in \mathbb{R}^{m \times K}$ is obtained. The prediction $\hat{\bm{y}} \in \mathbb{R}^m$ is obtained considering the outputs of $l_1,...,l_K$, and $\bm{P} \in \mathbb{R}^{m \times K}$.\\
Since Bayesian networks were created using six different methods, six prediction values are obtained. The accuracy of the six networks are compared to determine which method to use.

\item 
The construction method that achieves the highest accuracy among the six predicted values is selected. With the selected method, the accuracies obtained when learning by determining the number of divisions dynamically using X-means \cite{pelleg} and when learning when the number of divisions $K$($K$ = 2,3,4,5,6,7) is fixed in advance are compared. The most accurate division method is selected as the proposed method.
\end{enumerate}

\subsection{Proposed Method}
As described in the previous section, the proposed method associates a Bayesian network with multiple deep learners after training each learner to determine the learner to which data are to be input. A Bayesian classifier is applied to solve classification problems using Bayes' theorem. In a previous study \cite{JDAIP2017}, we adopted a naive Bayes classifier. A naive Bayes classifier assumes conditional independence between feature quantities and functions with a Bayesian network with the simplest possible structure. However, because this classifier is limited to the simplest structure, it is possible that the relationship between the nodes may not be expressed sufficiently. Therefore, in this study, the Bayesian network that constructed the network structure from the data is used as the Bayes classifier, without limiting the structure of the network.

%%%
\subsubsection{\textbf{Clustering and Training of Deep Learners}}
Let $\bm{S} \in \mathbb{R}^{n \times d}$ be the training data and $\bm{y} \in \{0,1\}^n$ be the label data. 
First, training dataset $\bm {S}$ is divided into $K$ clusters. 
Training data for each cluster, corresponding label data $(\bm{S}_c,\bm{y}_c )_{c=1}^{K}$, and a vector $\bm{c} \in \{1,2,...,K \}^n$ that stores the cluster index to which each row of $\bm{S}$ belongs are obtained. When using K-mean clustering, the number of clusters must be specified in advance. Here, the number of clusters was determined via X-means\cite{pelleg}. $K$ deep learners  $l_1,...,l_K$ are trained with $K$ clustered data $(\bm{S}_c,\bm{y}_c )_{c = 1}^{K}$. These learners are evaluated using the softmax activation function in the final layer. Here, the softmax function outputs a two-dimensional vector.

\subsubsection{\textbf{Training the Bayesian network}}
The Bayesian network is trained using data on $(\bm{S} ,\bm{c})$. 
When $d$ dimensional data $(x_1,...,x_d)^{\mathrm{T}}$ are given, we construct a classifier that probabilistically associates which learner among $K$ learners $(x_1,...,x_d)^{\mathrm{T}}$ should receive an input.
Let $\bm{X} = (X_1,X_2,...,X_d,X_{d+1})$ be a collection of $d +1$ random variables. $X_1, ..., X_d$ are the random variables corresponding to the input, and $X_{d+1}$ is the random variable corresponding to the selected learner. $X_{d+1}$ corresponds to $X_{l}$ in Figure \ref{fig:methodA}. 
Each variable adopts a single value from the $r_i$ states set $\{ 0,1,...,r_{i-1} \}  \; (i = 1,...,d+1)$. 
Given the structure $Bs$ of the Bayesian network and representing the parent node of variable $i$ as $pa(X_i)$, the joint probability distribution can be expressed as follows. 

\begin{eqnarray}
\scalebox{1.0}{$\displaystyle
P(X_1,X_2,...,X_{d+1}|B_s) = \prod_{i = 1}^{d+1} P(X_i|pa(X_i),B_s)
$}
\label{eq:0}
\end{eqnarray}

Note, it is assumed that the data are complete and there are no missing values.

When predicting from observed values or test data, marginalization is performed based on the Bayes' theorem, and the estimated conditional probability. The algorithm for the probability propagation method executes marginalization efficiently.

The Bayesian network is treated for (i) structure determination, (ii) conditional probability estimation, and (iii) marginalization using the probability propagation method. 
Training the Bayesian network is completed in steps (i) and (ii). 
The Bayesian network can handle continuous values by assuming a normal distribution to the conditional probability between nodes \cite{gaussian_net}. 
In this study, we construct a Bayesian network comprising continuous values; however, for simplicity, we assume that each variable takes a discrete value.

(i)\textbf{ Structure determination}.
It is known that the number of structural combinations in a Bayesian network increases explosively with respect to the number of nodes. 
For example, when the number of nodes is two, the number of combinations is three, when the number of nodes is three, the number of combinations is 25, and when the number of nodes is 5, the number of combinations is 29,281. 
Therefore, finding the structure of the Bayesian network using a full search is limited by the number of nodes. The structure of the Bayesian network is determined from prior knowledge of variables, information quantity criteria, and prediction distribution. In this study, the structure of the Bayesian network is determined heuristically from the data based on the information criterion. We constructed a Bayesian network with the hill climbing algorithm and tabu search, using log likelihood, AIC, and BIC as the information criteria.
 
Depending on the information criterion and the structure search algorithm, we employed the following six methods to construct Bayesian network: i) the hill climbing algorithm with log likelihood as a reference; ii) the hill climbing algorithm with AIC as a reference; iii) the hill climbing algorithm with BIC as a reference; iv) tabu search with log likelihood as a reference; v) tabu search with AIC as a reference; and vi) tabu search with BIC as a reference.
The accuracy of the Bayesian networks constructed using these six methods was compared to determine which method to use.
Let $I_B$ be the information criteria, which is expressed as follows, denoted as Equation(\ref{eq:1}).

\begin{eqnarray}
\scalebox{1.0}{$\displaystyle
I_B = -2\sum_{i=1}^{d+1} \sum_{j=1}^{q_i} \sum_{k=0}^{r_i -1} n_{ijk}\ln{\frac{n_{ijk}}{n_{ij} }} + c_B\sum_{i = 1}^{d+1} q_i(r_i -1)
$}
\label{eq:1}
\end{eqnarray}
Here, $n'_{ij} = \sum_{k=0}^{r_i-1}n'_{ijk}, n_ {ij} = \sum_{k=0}^{r_i-1}n_{ijk}$.

Let the number of states that the random variable $X_i$ can assume be $r_i$, and the total state of the parent variables of $X_i$ be $q_i$. Here, $n_{ijk}$ indicates the number of data items that becomes $X_i = k$ when $pa(X_i) = j$.

The first term in Equation(\ref{eq:1}) represents the log likelihood, and the second term is the number of parameters. By appropriately selecting the value of $c_B$, we can construct the information quantity criterion. 
For example, when $c_B = 2$, Equation(\ref{eq:1}) becomes the AIC and when $c_B = \log(d+1)$ it becomes the BIC. 
Even in the Bayesian network, we need to find a structure that optimizes these information criteria. The information quantity criterion is often used as an index of a heuristic search method, and in this study the information quantity criterion was used as an index when searching the structure via the hill climbing algorithm or tabu search.

(ii)\textbf{ Estimating conditional probability}.
Estimating the conditional probability $P(X_i | pa (X_j))$ of each node from the data is necessary. 
Let $\theta_{ijk} = P(X_i = k | pa(Xi) = j)$ be the parameter to be estimated. 
$\Theta_{B_s} = (\theta_{ijk})   (i = 1,...,d+1, j = 1,...,q_i, k = 0,...,r_{i-1})$

At this instance, the likelihood $L(\Theta_{B_s} | \bm{S}) $ with the given data $\bm{S}$ is expressed as follows.
\begin{eqnarray}
\scalebox{1.0}{$\displaystyle
L(\Theta_{B_s}|\bm{S}) \propto \prod_{i = 1}^{d+1}\prod_{j = 1}^{q_i}\prod_{k = 0}^{r_i - 1} \theta_{ijk}^{n_{ijk}}
$}
\label{eq:2}
\end{eqnarray}

Because the likelihood distribution of the Bayesian network follows a multinomial distribution, a Dirichlet distribution is set as a conjugate prior distribution.
\begin{eqnarray}
\scalebox{1.0}{$\displaystyle
P(\Theta_{B_s}) = \prod_{i = 1}^{d+1}\prod_{j = 1}^{q_i} \prod_{k = 0}^{r_i - 1}\frac{\Gamma(n'_{ij})}{\Gamma(n'_{ijk})}\theta_{ijk}^{n'_{ijk}-1}
$}
\label{eq:3}
\end{eqnarray}
Here, $\Gamma()$ represents a gamma function and $n'_{ijk}$ represents the hyperparameter of the prior distribution corresponding to $n_{ijk}$, $n'_{ijk} > 0\quad (k = 0, ..., r_i - 1)$.

Because the posterior distribution is the product of the prior distribution and likelihood, the posterior distribution is given as follows.
\begin{eqnarray}
\scalebox{1.0}{$\displaystyle
P(\Theta_{B_s}|S) \propto \prod_{i = 1}^{d+1}\prod_{j = 1}^{q_i}\prod_{k = 0}^{r_i - 1} \theta_{ijk}^{n'_{ijk} + n_{ijk} -1}
$}
\label{eq:4}
\end{eqnarray}

The estimated value of conditional probability $\hat{\theta_{ijk}}$ can be obtained from expression (\ref {eq:4}).
The maximum a posteriori estimate $\theta_{MAP}$ of expression (\ref{eq:4}) is given as follows.
\begin{eqnarray}
\scalebox{1.0}{$\displaystyle
\theta_{MAP} = \frac{n'_{ijk}+n_{ijk}-1}{n'_{ij}+n_{ij}-r_{i}}  \quad  (k = 0,...,r_{i}-2)
$}
\label{eq:5}
\end{eqnarray}

\subsubsection{\textbf{Prediction}}
Let $\bm{S}_{test} = (\bm{s}_1,...,\bm{s}_m)^{\mathrm{T}} \in \mathbb{R}^{m \times d}$ be the test data.

When the conditional probability is estimated for each node, prediction can be performed by calculating the posterior probabilities of the nodes of interest from specific observation values or test data and comparing them. 
Using the probability propagation method, peripheralization can be performed efficiently. 
$X_1,...,X_d$ is a random variable representing each input variable, and $X_{l}$ is a random variable indicating which learner the input is associated with.

$\bm{S}_{test}$ are inputs to the Bayesian network and the posterior probability $\bm{P} \in \mathbb{R}^{m \times K}$ is obtained.
\begin{eqnarray}
\scalebox{1.0}{$\displaystyle
(\bm{P})_{bc} = P(X_{l} = c |\bm{s}_b) \quad (b = 1,...,m, c = 1,...,K)
$}
\label{eq:6}
\end{eqnarray}

Let $h$ be a threshold. $\hat{\bm{P}}$ is defined by Equation(\ref{eq:7}).
In this study, $h$ is 0.001.
\begin{eqnarray}
\scalebox{1.0}{$\displaystyle
(\hat{\bm{P}})_{bc} = \begin{cases}
(\bm{P})_{bc} & ((\bm{P})_{bc} \geq h) \\
0 &  ((\bm{P})_{bc} < h)
\end{cases}
$}
\label{eq:7}
\end{eqnarray}

The predicted value $\hat{y}_b$ is obtained as shown in Equation (\ref{eq:8}) from the outputs from the $K$ deep learners $l_1(\bm{s}_b),...,l_K(\bm{s}_b)$, which output two-dimensional vectors and the probabilities $\bm{P}$ obtained from the Bayesian network.
\begin{eqnarray}
\scalebox{1.0}{$\displaystyle
\hat{y}_b = \argmax \left( \sum^K_{c = 1} \hat{(\bm{P})}_{bc}l_c(\bm{s}_b) \right)
$}
\label{eq:8}
\end{eqnarray}
The Bayesian network was created using six methods; thus, six prediction values can be obtained. The accuracy of the Bayesian networks constructed using these six methods is compared to determine which method to use.

\subsubsection{\textbf{Determining the number of clusters $K$ }}
In the proposed method, training dataset $\bm {S}$ is divided into $K$ clusters via K-means clustering. The number of clusters must be specified in advance. 
There are two ways to determine division number $K$, i.e., by determining $K$ dynamically according to the data, as proposed by Pelleg et al. \cite{pelleg}, and by determining $K$ in advance.

The structure of the Bayesian network is determined by the method discussed in the previous section 2.2.2 and 2.2.3. The accuracy when learning is performed by dynamically determining the number of divisions using X-means \cite{pelleg} and the accuracy when learning is performed with the number of clusters $K$ is fixed ($K$ = 2,3,4,5,6,7) in advance are compared, and the most accurate division method is selected as the proposed method.

\section{Experiments}
Numerical experiments were performed to evaluate the proposed method.

We compared the following five methods. 
The first method involves a single deep learner; data are not divided, and multiple learners are not trained. 
The second method divides the training data via K-means and uses the K-means cluster centers to associate the test data with multiple learners. 
The third method is the Deep MoE method proposed by Eigen et al. \cite{eigen2014}, and the fourth is the Hard MoE model proposed by Gross et al. \cite{sam2017}. The model and training details of gater $T$ with $L$ layers and experts $H_c (c = 1,...,K)$ follow the method Gross et al. proposed \cite{sam2017}. 
In this method, PCA is performed on the output of the $L-1$ layer of $T$, and K-means clustering is performed with $K$ centroids. 
Gater $T$ associates the inputs with $H_c$. 
However, the data used in the experiment are different; thus, the structures and parameters of gater $T$ and experts $H_c$ are matched to those used in this experiment. The fifth method is our previous method \cite{JDAIP2017}. In a previous study \cite{JDAIP2017}, we divided training data via K-means and used a naive Bayes classifier to associate the test data with multiple learners. In that method, the number of clusters $K$ is determined dynamically according to the data.
To compare the proposed method, the experiments were performed using the same dataset.

\subsection{Data}
In this section, we explain the data used in the experiment.

We used financial time-series data of six indicators to predict deviations from the average return of the Nikkei Average. The data comprised the closing prices of the daily data of the Nikkei Stock Average, NY Dow, NASDAQ, S\&P 500, FTSE 100, and DAX from January 1, 2000 to December 31, 2014. These data were obtained from Yahoo! Finance, the Federal Reserve Bank of St. Louis. The NY Dow, NASDAQ, and S\&P 500 are American stock price indexes. The FTSE 100 and DAX are European stock indexes.

Typically, stock exchanges are closed on national holidays. Consequently, data may be not be available. In this case, an indicator with no data was assumed to be unchanged from the previous day, and the index value of the previous day was adopted. Generally, raw financial time-series data tend to have strong non-stationarity; thus, appropriate deformation is required. In this study, six financial stock data were transformed into returns.

Here, let $\bm{p}(t)  \;(0 \leq t \leq T)$ be the time-series data. 
According to Equation (\ref{eq:return}), $\bm{p}(t - 1)$ and $\bm{p}(t)$ are transformed to return $\bm{r}(t)$.

\begin{eqnarray}
\scalebox{1.0}{$\displaystyle
\bm{r}(t) = \{ \log \bm{p}(t) - \log \bm{p}(t - 1) \} \times 100
$}
\label{eq:return}
\end{eqnarray}
The return $\bm{r}(t) \; (0 \leq t \leq T) $ is obtained by shifting the data one period at a time.

Furthermore, the Dickey Fuller test was conducted to confirm the continuity of return $\bm{r}(t)$, where the null hypothesis was that unit roots exist. 
In contrast, the alternative hypothesis was that the tested time-series data represent a stationary process. From the results of the Dickey Fuller test, we assumed that return $\bm{r}(t)$ is a stationary time-series because the null hypothesis was rejected at a dominance level of 5\%. We predicted divergence from the average return of the Nikkei Stock Average in two classes. Based on the stationarity assumption, the average return was constant for all time periods.

\subsection{Experimental Results}
This section presents the results of the numerical experiments.

We predicted the divergence from the average return of the Nikkei Stock Average in two classes. Specifically, based on return $\bm{r}(t)$ in period $t$, we predicted whether return $r(t + 1)$ of the Nikkei Average stock price in period $t+1$ was greater or less than the average return. Here, the total number of data was 3912. 
We used 3652 data from 2000 to 2013 as training data and 260 data from 2014 as test data. The training data were$\bm{S} \in \mathbb{R}^{3652 \times 6}$, and the label data for the training data were $\bm{y} \in \{1,-1\}^{3652}$. The test data were $\bm{S}_{test} \in \mathbb{R}^{260 \times 6}$, and the label data for the test data were $\bm{y}_{test} \in  \{1,-1\}^{260}$.

Three types of deep learners, i.e., DNN, RNN, and LSTM, were used for multiple learners. All deep learners comprised two hidden layers, and each hidden layer had six units.

The hidden layer used ReLU as the activation function and bias. The softmax activation function was used to evaluate the final layer, and a two-dimensional vector was output.

The weights of the deep learners were learned over 100 iterations; the dropout rate was 0.2. The minibatch size was one-fifth of the data clustered to each learner.

To train the proposed model, we trained deep learners and a Bayesian network. We trained multiple deep learners using the TensorFlow software library with an NVDIA Tesla K80 GPUs (with 12 GB and 13 GB RAM). We trained the Bayesian network using the bnlearn package \cite{marco2010} with a 3.1 GHz Intel Core i7 with 16 GB RAM.

We performed each experiment 100 times to determine the mean, standard error, and computation time.

\subsubsection{\textbf{Multiple DNN learners}}
\begin{figure}[!h]
  \begin{center}
    \includegraphics[clip,width=8.0cm]{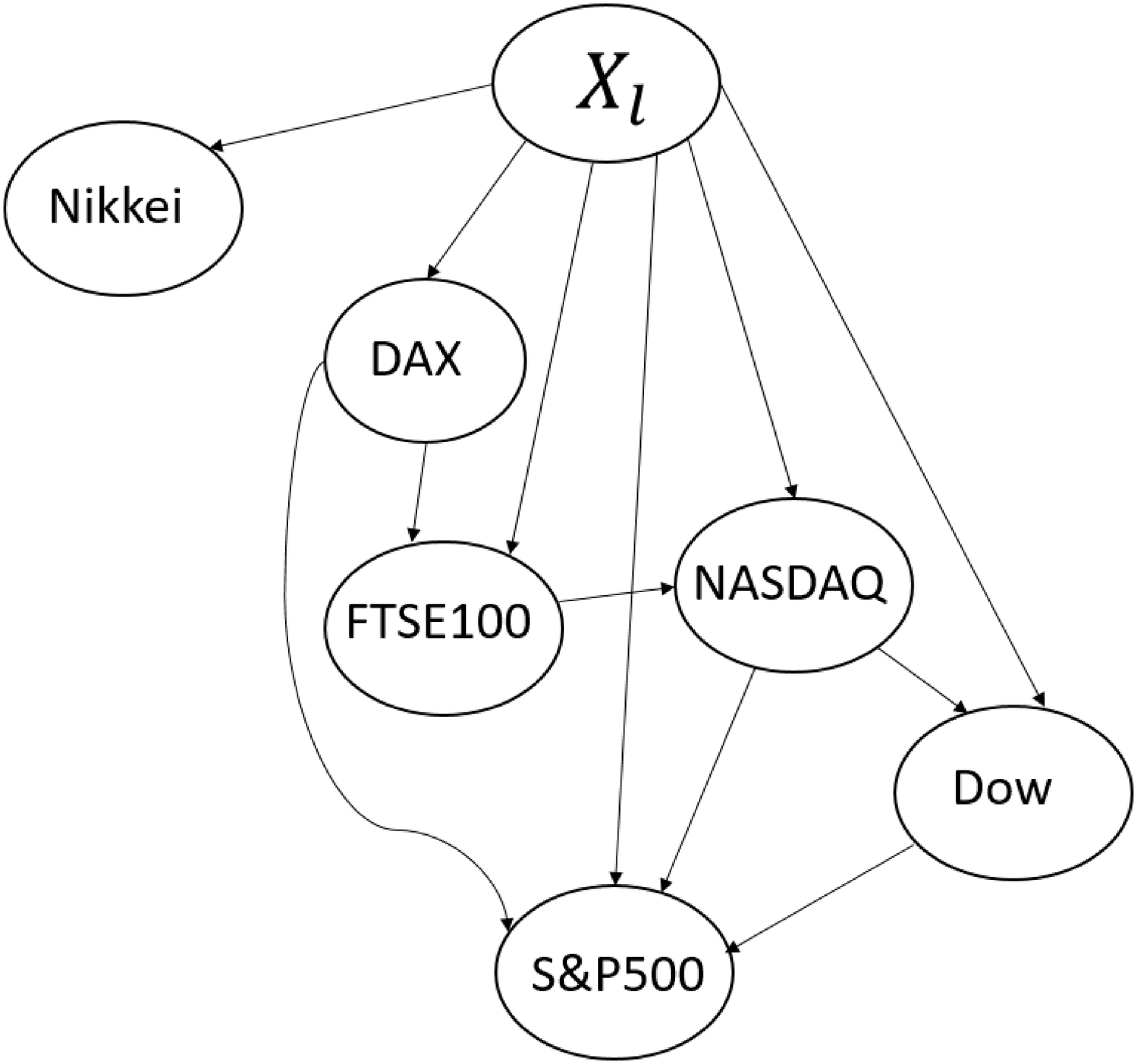}
    \caption{Bayesian network constructed by hill climbing algorithm with BIC when the highest accuracy and f-values were obtained by the proposed method. Data were divided into 6 clusters with K-means. Bayesian network and 6 DNN learners corresponding to each cluster were trained. This Bayesian network maps data to one of six DNN learners.}
    \label{fig:bn_dnn}
  \end{center}
\end{figure}

Table 1 presents the average accuracy and f-values for each experiment with multiple DNN learners. 
The method to construct the Bayesian network that demonstrated the highest accuracy was the hill climbing algorithm with BIC. 
Table 2 presents the average accuracy and f-values when the number of clusters $K$ was changed and the network was constructed via the hill climbing algorithm with BIC. 
Here, the highest accuracy was achieved when the number of divisions was fixed at six. 
Figure \ref{fig:bn_dnn} shows the Bayesian network when the highest accuracy and f-values were obtained by the proposed method. 
Table 3 shows a comparison of the proposed method and five other methods. 
With the Deep MoE model \cite{eigen2014}, we constructed two layers of Deep MoE. The first layer comprised three DNNs (expert networks) with six, five, and five units from the input and one DNN (gate network) with six, five, and three units. The second layer comprised three DNNs (expert networks) with five, three, and two units from the input and one DNN (gate network) with five, five, three, and two units. For the Hard Mixture Experts model \cite{sam2017}, the number of clusters was $K = 6$.

Table 4 shows the average and maximum computation time for multiple learner cases. 
The average computation time shows the average and standard deviation of the time required to train all learners. 
In contrast, the maximum computation time with multiple learners shows the average and standard deviation of the computation time of the learner requiring the maximum calculation time with multiple learners. 
When the prediction was performed with a single learner, the maximum computation time with multiple learners represents the maximum calculation time of 100 experiments.

\begin{table*}[!h]
    \begin{center}
  \caption{Accuracy and F-values when dividing data by X-means and training multiple DNN learners}
\scalebox{1.0}{
  \begin{tabular}{ccc}\hline
    \shortstack{Method to construct \\ Bayesian network} &  Accuracy & F-value  \\ \hline 
	Hill climbing algorithm\\with log likelihood & 0.6744 $\pm$ 0.0101 &  0.6546 $\pm$ 0.0129  \\ 
	Hill climbing algorithm\\with AIC & 0.6724 $\pm$ 0.0091 & 0.6532 $\pm$ 0.0117 \\
	Hill climbing algorithm\\with BIC &{\bf 0.6765 $\pm$ 0.0125} & {\bf 0.6564 $\pm$ 0.0195} \\
	Tabu search with log likelihood & 0.6738 $\pm$ 0.0012 & 0.6555 $\pm$ 0.0190     \\ 
     Tabu search with AIC  &  0.6746 $\pm$ 0.0118 & 0.6553 $\pm$ 0.0178    \\ 
     Tabu search with BIC  &  0.6743 $\pm$ 0.0131 & 0.6540 $\pm$ 0.0204    \\     \hline
     \end{tabular}
}
    \end{center}
    \label{tab:result_dnn}
 \end{table*}

\begin{table*}[!h]
    \begin{center}
  \caption{Accuracy and F-values when constructing Bayesian network with hill climbing algorithm with BIC}
\scalebox{1.0}{
  \begin{tabular}{ccc}\hline
    \shortstack{Number  of clusters $K$} &  Accuracy & F-value  \\ \hline 
	$K$ = 2 & 0.6698 $\pm$ 0.0156 &  0.6433 $\pm$ 0.0305  \\ 
	$K$ = 3 & 0.6781 $\pm$ 0.0147 &  {\bf 0.6692 $\pm$ 0.0151}  \\ 
	$K$ = 4 & 0.6746 $\pm$ 0.0103 &  0.6599 $\pm$ 0.0127  \\ 
	$K$ = 5 & 0.6751 $\pm$ 0.0106 &  0.6633 $\pm$ 0.0131  \\ 
	$K$ = 6 & {\bf 0.6810 $\pm$ 0.0126} &  0.6689 $\pm$ 0.0140  \\ 
	$K$ = 7 & 0.6758 $\pm$ 0.0139 &  0.6652 $\pm$ 0.0182  \\ 
	Dynamically Determination \\with X-means\cite{pelleg} & 0.6765 $\pm$ 0.0125 & 0.6564 $\pm$ 0.0195 \\ \hline
     \end{tabular}
}
    \end{center}
    \label{tab:result_dnn}
 \end{table*}

\begin{table*}[!h]
    \begin{center}
  \caption{Accuracy and F-values of proposed and other methods}
\scalebox{1.0}{
  \begin{tabular}{ccc}\hline
    \shortstack{Classifier} &  Accuracy & F-value  \\ \hline 
     Single DNN  &  0.6723 $\pm$ 0.0104 & 0.6669 $\pm$ 0.0108  \\ 
	K-mean to Associate\\Multiple Learners with Data& 0.6729 $\pm$ 0.0121&0.6596 $\pm$ 0.0135  \\ 
	Deep MoE\cite{eigen2014} &  0.6580   $\pm$ 0.0257 & 0.6336 $\pm$ 0.0535 \\
	Hard Mixture Experts\cite{sam2017} & 0.5732 $\pm$ 0.0886 & 0.5211 $\pm$ 0.1057 \\
	Naive Bayes\cite{JDAIP2017} &  0.6787 $\pm$ 0.0093 & 0.6544 $\pm$ 0.0115     \\ 
     Proposed method($K$=6)  &  {\bf 0.6810 $\pm$ 0.0126} & {\bf 0.6689 $\pm$ 0.0140}   \\     \hline
     \end{tabular}
}
    \end{center}
    \label{tab:result_dnn}
 \end{table*}

\begin{table*}[!h]
    \begin{center}
  \caption{Computation Time}
\scalebox{1.0}{
  \begin{tabular}{ccc}\hline
    \shortstack{Classifier} & Computation Time[s]  & \shortstack{\\Maximum computation time \\ in multiple learners[s]}  \\ \hline 
     Single DNN  &  3.0643 $\pm$ 0.1186 & 3.5357  \\ 
	K-mean to Associate\\Multiple Learners with Data& 19.8858 $\pm$ 0.1250 & 3.5430 $\pm$ 0.0885  \\ 
	Deep MoE\cite{eigen2014} & 312.2613 $\pm$ 33.1540 & 3.6448 $\pm$ 0.4720 \\
	Hard Mixture Experts\cite{sam2017} & 22.7472 $\pm$ 0.9818 & 3.5886 $\pm$ 0.0742 \\
	Naive Bayes\cite{JDAIP2017} &  44.0864 $\pm$ 8.7180 & 4.3812 $\pm$ 0.4382     \\ 
     Proposed method($K$=6)  &    26.3775 $\pm$ 0.1811 &  3.5295 $\pm$ 0.0725   \\     \hline
     \end{tabular}
}
    \end{center}
    \label{tab:result_dnn}
 \end{table*}

\subsubsection{\textbf{Multiple RNN learners}}
\begin{figure}[!h]
  \begin{center}
    \includegraphics[clip,width=8.0cm]{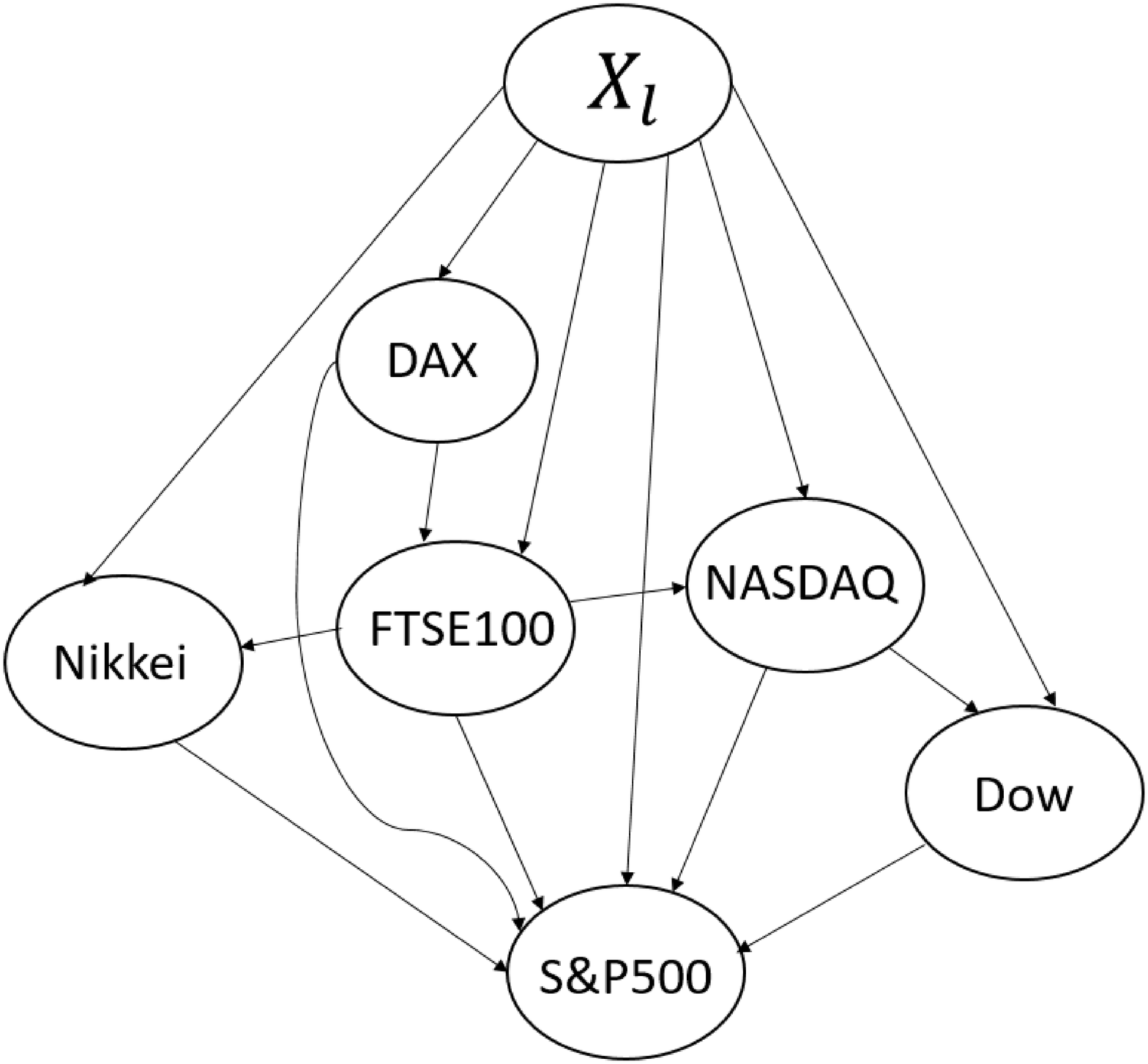}
    \caption{Bayesian network constructed by hill climbing algorithm with AIC when highest accuracy and f-values were obtained by the proposed method.  Data were divided into 6 clusters with K-means. Bayesian network and 6 RNN learners corresponding to each cluster were trained. This Bayesian network maps data to one of six RNN learners.}
    \label{fig:bn_rnn}
  \end{center}
\end{figure}

Table 5 shows the average accuracy and f-values obtained with multiple RNN learners. 
Here, the method to construct the Bayesian network that demonstrated the highest accuracy was the hill climbing algorithm with AIC. 
Table 6 presents the average accuracy and f-values when the number of clusters $K$ was changed and the network was constructed using the hill climbing algorithm with AIC. 
As can be seen, the highest accuracy was obtained with six divisions. 
Figure \ref{fig:bn_rnn} shows the Bayesian network when the highest accuracy and f-values were obtained by the proposed method, and Table 7 compares the proposed method and five other methods. With Deep MoE \cite{eigen2014}, we constructed two Deep MoE layers. The first layer comprised three RNNs (expert networks) with six, five, and five units from the input and one RNN (gate network) with six, five, and three units. The second layer comprised three RNNs (expert networks) having five,three, and two units from the input and one RNN (gate network) having five, five, three, and two units. The number of clusters is $K = 6$ in Hard Mixture Experts model \cite{sam2017}.
Table 8 shows the average and maximum computation time for multiple learner cases.

\begin{table*}[!h]
    \begin{center}
  \caption{Accuracy and F-values when dividing data by X-means and training multiple RNN learners}
\scalebox{1.0}{
  \begin{tabular}{ccc}\hline
    \shortstack{\\Method to construct \\ Bayesian network} &  Accuracy & F-value  \\ \hline 
	Hill climbing algorithm\\with log likelihood & 0.6669 $\pm$ 0.0117 &  0.6449 $\pm$ 0.0154  \\ 
	Hill climbing algorithm\\with AIC & {\bf 0.6692 $\pm$ 0.0117} & {\bf 0.6475 $\pm$ 0.0152} \\
	Hill climbing algorithm\\with BIC & 0.6672 $\pm$ 0.0114 &  0.6452 $\pm$ 0.0143 \\
	Tabu search with log likelihood & 0.6666 $\pm$ 0.0120 & 0.6445 $\pm$ 0.01670     \\ 
     Tabu search with AIC  &  0.6662 $\pm$ 0.0094 & 0.6449 $\pm$ 0.0127    \\ 
     Tabu search with BIC  &  0.6676 $\pm$ 0.0131 & 0.6465 $\pm$ 0.0174    \\     \hline
     \end{tabular}
}
    \end{center}
    \label{tab:result_dnn}
 \end{table*}

\begin{table*}[!h]
    \begin{center}
  \caption{Accuracy and F-values when constructing Bayesian network with hill climbing algorithm with AIC}
\scalebox{1.0}{
  \begin{tabular}{ccc}\hline
    \shortstack{Number of clusters $K$} &  Accuracy & F-value  \\ \hline 
	$K$ = 2 & 0.6605 $\pm$ 0.0160 &  0.6327 $\pm$ 0.0308  \\ 
	$K$ = 3 & 0.6695 $\pm$ 0.0155 &  0.6552 $\pm$ 0.0190  \\ 
	$K$ = 4 & 0.6668 $\pm$ 0.0132 &  0.6504 $\pm$ 0.0161  \\ 
	$K$ = 5 & 0.6729 $\pm$ 0.0184 &  0.6595 $\pm$ 0.0237  \\ 
	$K$ = 6 & {\bf 0.6798 $\pm$ 0.0144} & {\bf 0.6630 $\pm$ 0.0216}  \\ 
	$K$ = 7 & 0.6667 $\pm$ 0.0154 &  0.6528 $\pm$ 0.0207  \\ 
	Dynamically Determination \\with X-means\cite{pelleg} & 0.6692 $\pm$  0.0117 & 0.6475 $\pm$ 0.0152 \\ \hline
     \end{tabular}
}
    \end{center}
    \label{tab:result_dnn}
 \end{table*}

\begin{table*}[!h]
    \begin{center}
  \caption{Accuracy and F-values of proposed and other methods}
\scalebox{1.0}{
  \begin{tabular}{ccc}\hline
    \shortstack{Classifier} &  Accuracy & F-value  \\ \hline 
     Single RNN  &  0.6683 $\pm$ 0.0126 & 0.6568 $\pm$ 0.0151  \\ 
	K-mean to Associate\\Multiple Learners with Data& 0.6690 $\pm$ 0.0132&0.6508 $\pm$ 0.0181  \\ 
	Deep MoE\cite{eigen2014} & 0.6446   $\pm$ 0.0317 & 0.6155 $\pm$ 0.0741 \\
	Hard Mixture Experts\cite{sam2017} & 0.5604 $\pm$ 0.0793 & 0.5005 $\pm$ 0.1002 \\
	Naive Bayes\cite{JDAIP2017} &  0.6780 $\pm$ 0.0113 & 0.6541 $\pm$ 0.0125     \\ 
     Proposed method($K$=6)  &  {\bf 0.6798 $\pm$ 0.0144} & {\bf 0.6630 $\pm$ 0.0216}   \\     \hline
     \end{tabular}
}
    \end{center}
    \label{tab:result_dnn}
 \end{table*}

\begin{table*}[!h]
    \begin{center}
  \caption{Computation Time}
\scalebox{1.0}{
  \begin{tabular}{ccc}\hline
    \shortstack{Classifier} & Computation Time[s]  & \shortstack{\\Maximum computation time \\in multiple learners[s]}   \\ \hline 
     Single RNN  &  9.702 $\pm$ 0.1024 & 9.9230  \\ 
	K-mean to Associate\\Multiple Learners with Data& 60.6614 $\pm$ 1.3407&10.9438 $\pm$ 0.4038  \\ 
	Deep MoE\cite{eigen2014} & 989.7782   $\pm$ 25.1792 & 10.6493 $\pm$ 0.3675 \\
	Hard Mixture Experts\cite{sam2017} & 71.4432 $\pm$ 4.1851 & 11.2840 $\pm$ 0.5378 \\
	Naive Bayes\cite{JDAIP2017} &  122.3491 $\pm$ 23.2997 & 12.2384 $\pm$ 1.2961     \\ 
     Proposed method($K$=6)  &  64.0060 $\pm$ 0.4854 &  9.8388 $\pm$ 0.1346  \\     \hline
     \end{tabular}
}
    \end{center}
    \label{tab:result_dnn}
 \end{table*}

\subsubsection{\textbf{Multiple LSTM learners}}
\begin{figure}[!h]
  \begin{center}
    \includegraphics[clip,width=8.0cm]{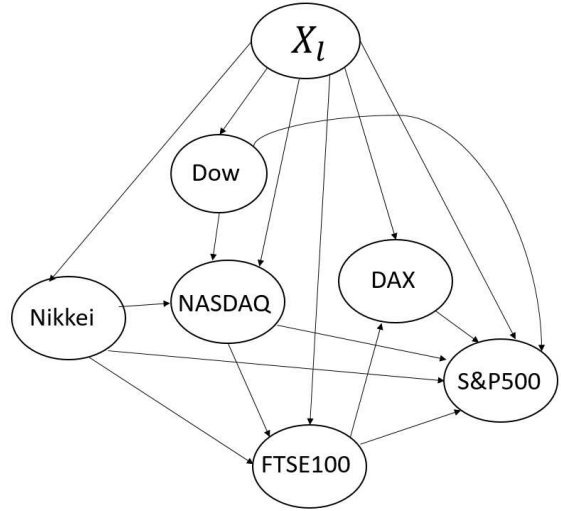}
    \caption{Bayesian network constructed by tabu search with AIC when the highest accuracy and f-values were obtained by the proposed method. Data were divided into 6 clusters with K-means. Bayesian network and 6 LSTM learners corresponding to each cluster were trained. This Bayesian network maps data to one of six LSTM learners.}
    \label{fig:bn_lstm}
  \end{center}
\end{figure}

Table 9 presents the average accuracy and f-values obtained with multiple LSTM learners. The Bayesian network construction method that demonstrated the highest accuracy was tabu search with AIC. Table 10 presents the average accuracy and f-values when the number of clusters $K$ was changed and the network was constructed via tabu search with AIC. 
As shown, the highest accuracy was obtained with six divisions. 
Figure \ref{fig:bn_lstm} shows the Bayesian network when the highest accuracy and f-values were obtained by the proposed method, and Table 11 compares the results of the proposed and five other methods. 
With Deep MoE \cite{eigen2014}, we constructed two Deep MoE layers. The first layer comprised two DNNs (expert networks) with six, six, and five units from the input and one DNN (gate network) with six, six, and two units. The second layer comprised three LSTMs (expert networks) with five, five, and two units from the input and one DNN (gate network) with five, three, and two units. Here, the number of clusters was $K = 6$ in Hard Mixture Experts model \cite{sam2017}.
Table 12 shows the average and maximum computation time for multiple learner cases.

\begin{table*}[!h]
    \begin{center}
  \caption{Accuracy and F-values when dividing data by X-means and training multiple LSTM learners}
\scalebox{1.0}{
  \begin{tabular}{ccc}\hline
    \shortstack{\\Method to construct \\ Bayesian network} &  Accuracy & F-value  \\ \hline 
	Hill climbing algorithm\\with log likelihood & 0.6677 $\pm$ 0.0087 &  0.6484 $\pm$ 0.0118  \\ 
	Hill climbing algorithm\\with AIC & 0.6681 $\pm$ 0.0074 & 0.6471 $\pm$ 0.0111 \\
	Hill climbing algorithm\\with BIC & 0.6683 $\pm$ 0.0078 & {\bf 0.6490 $\pm$ 0.0107} \\
	Tabu search with log likelihood & 0.6662 $\pm$ 0.0087 & 0.6459 $\pm$ 0.0113    \\ 
     Tabu search with AIC  & {\bf 0.6688 $\pm$ 0.0089} & 0.6484 $\pm$ 0.0116    \\ 
     Tabu search with BIC  &  0.6669 $\pm$ 0.0070 & 0.6464 $\pm$ 0.0091    \\     \hline
     \end{tabular}
}
    \end{center}
    \label{tab:result_dnn}
 \end{table*}

\begin{table*}[!h]
    \begin{center}
  \caption{Accuracy and F-values when constructing Bayesian network with tabu search with AIC}
\scalebox{1.0}{
  \begin{tabular}{ccc}\hline
    \shortstack{Number of clusters $K$} &  Accuracy & F-value  \\ \hline 
	$K$ = 2 & 0.6602 $\pm$ 0.0195 &  0.6370 $\pm$ 0.0375  \\ 
	$K$ = 3 & 0.6847 $\pm$ 0.0116 &  0.6724 $\pm$ 0.0120  \\ 
	$K$ = 4 & 0.6754 $\pm$ 0.0112 &  0.6603 $\pm$ 0.0131  \\ 
	$K$ = 5 & 0.6800 $\pm$ 0.0087 &  0.6659 $\pm$ 0.0101  \\ 
	$K$ = 6 & {\bf 0.6902 $\pm$ 0.0109} & {\bf 0.6749 $\pm$ 0.0121}  \\ 
	$K$ = 7 & 0.6765 $\pm$ 0.0101 &  0.6633 $\pm$ 0.0113  \\ 
	Dynamically Determination \\with X-means\cite{pelleg} & 0.6688 $\pm$ 0.0089 & 0.6484 $\pm$ 0.0116 \\ \hline
     \end{tabular}
}
    \end{center}
    \label{tab:result_dnn}
 \end{table*}

\begin{table*}[!h]
    \begin{center}
  \caption{Accuracy and F-values of proposed and other methods}
\scalebox{1.0}{
  \begin{tabular}{ccc}\hline
    \shortstack{Classifier} &  Accuracy & F-value  \\ \hline 
     Single LSTM  &  0.6679 $\pm$ 0.0113 & 0.6574 $\pm$ 0.0115  \\ 
	K-mean to Associate\\Multiple Learners with Data& 0.6755 $\pm$ 0.0071&0.6554 $\pm$ 0.0084  \\ 
	Deep MoE\cite{eigen2014} & 0.5841 $\pm$ 0.0378 & 0.4665 $\pm$ 0.0922 \\
	Hard Mixture Experts\cite{sam2017} & 0.5949 $\pm$ 0.0712 & 0.5272 $\pm$ 0.1064 \\
	Naive Bayes\cite{JDAIP2017} &  0.6769 $\pm$ 0.0082 & 0.6528 $\pm$ 0.0086     \\ 
     Proposed method($K$=6)  & {\bf 0.6902 $\pm$ 0.0109} & {\bf 0.6749 $\pm$ 0.0121}   \\     \hline
     \end{tabular}
}
    \end{center}
    \label{tab:result_dnn}
 \end{table*}

\begin{table*}[!h]
    \begin{center}
  \caption{Computation Time}
\scalebox{1.0}{
  \begin{tabular}{ccc}\hline
    \shortstack{Classifier} &  Computation Time[s]  & \shortstack{\\Maximum computation time \\in multiple learners[s]}   \\ \hline 
     Single LSTM  &  22.548 $\pm$ 0.1463 & 22.9499  \\ 
	K-mean to Associate\\Multiple Learners with Data& 129.5857$\pm$ 1.3524&23.2244$\pm$ 0.6274  \\ 
	Deep MoE\cite{eigen2014} & 582.3817  $\pm$ 20.7474 & 6.6540 $\pm$ 0.4992 \\
	Hard Mixture Experts\cite{sam2017} & 145.3129 $\pm$ 7.133 & 22.6399 $\pm$ 0.5058 \\
	Naive Bayes\cite{JDAIP2017} &  274.6611 $\pm$ 51.8509 & 27.0417 $\pm$ 2.8808     \\ 
     Proposed method($K$=6)  & 135.2146 $\pm$ 1.1884 &  22.6444 $\pm$ 0.5175   \\     \hline
     \end{tabular}
}
    \end{center}
    \label{tab:result_dnn}
 \end{table*}

\section{Discussion}
Here, we discuss the experimental results. In this study, we propose a data adaptive prediction method using a Bayesian network to associate multiple learners with data. A naive Bayes classifier was used in our previous study \cite{JDAIP2017}; however, in the current study a Bayesian network was used because it can express complicated relationships between variables. Thus, it is believed that better accuracy and f-values can be obtained using a Bayesian network. Generally, the proposed method demonstrated better results in terms of accuracy and f-values compared to adopting a single learner, K-means, Deep MoE \cite{eigen2014}, Hard Mixture Experts \cite{sam2017}, and our previous method \cite{JDAIP2017}.

We used naive Bayes classifiers in our previous studies \cite{JDAIP2017} to associate explanatory variables with multiple learners; however, in this study we use Bayesian networks. The proposed method has the following three significant characteristics. First, unlike the naive Bayes classifier, a Bayesian network can visualize the relationship between variables. Next, the accuracy obtained in this study is higher than that of the comparative method \cite{eigen2014}\cite{sam2017} and our previous study \cite{JDAIP2017}. Finally, since a Bayesian network is used, when a deficiency occurs in an explanatory variable, it is possible to estimate a deficient explanatory variable from other variables. This property is important for prediction in areas such as medicine where complete data is not always available. 
The proposed method can be implemented in other domains by changing the clustering method or different deep learners. The limitation of this study is that the objective variable must be a discrete value because it uses a Bayesian network.

Compared to naive Bayes approaches, with Bayesian networks, it is difficult to search for optimal structures rather than expressing complex structures. In this study, the structure of Bayesian network was searched heuristically. We used the hill climbing algorithm with BIC when the learner has multiple DNNs, hill climbing with AIC when the learners have multiple RNNs, and tabu search with AIC when the learners have multiple LSTMs. The network search methods differ because the types of multiple learners differ and because heuristic search methods are used. In our experiments, 100 trials were conducted to obtain the average value. However, an optimal network was not obtained in each trial. To address this problem, for prediction, we adopted multiple learners that indicate the probability that the data are associated with the given learners.

Tables 2, 6, and 10 show the accuracy when the data were divided with fixed $K$ and when the number of clusters $K$ was determined dynamically. 
The best accuracy was obtained with $K = 6$. If the number of clusters is too small, the data cannot be divided well. If there are too many clusters, over fitting becomes a problem. The reason why the highest performance was recorded when the number of clusters was 6 was that the data structure was divided successfully and generalization ability was obtained. When the number of clusters $K$ was determined dynamically, the accuracy was less than that of the $K = 6$ case; however,  it was not the lowest accuracy. Since X-means \cite{pelleg} can determine the value of $K$ from the data, it appears that there is an advantage whereby a suboptimal result can be obtained without searching for optimum $K$.

Figures \ref{fig:bn_dnn}, \ref{fig:bn_rnn} and \ref{fig:bn_lstm} show the optimal Bayesian network when multiple learners are DNNs, RNNS, and LSTMs. The stock markets corresponding to the nodes of these networks have edge connections that generally satisfy the time-series relationship that the stock market opens. On the other hand, there is an edge between DAX and S\&P 500, and Dow and S\&P 500 in the three networks. This means that these variables and edges are important in the prediction. 

Tables 3, 7, and 11 that show that the proposed method obtains higher accuracy and f-values than the other method. This is considered to indicate that the prediction that considers the probability of adopting multiple learners can be obtained robustly relative to the structure of the Bayesian network.

Tables 4, 8, and 12 show the computation time when using multiple learners. 
In some cases, it is confirmed that the proposed method's computation time might be less than that of our previous method \cite{JDAIP2017}, in which data and multiple learners are associated with K-means. In addition, when prediction is performed with multiple learners, the computation time required to train the learners tends to be longer than when using a single learner.

 If training multiple learners is performed via parallel computation, then the portion requiring the longest calculation time among multiple learners corresponds to the rate-limiting calculation step. Although parallelization was not employed in our experiments, it is highly probably that computation time can be reduced significantly by training multiple learners in parallel.

\section{Conclusion and Future Work}
In this paper, we have presented a general framework for the selection of multiple learners using a Bayesian network and proposed a method based on our previous study \cite{JDAIP2017} where a Bayesian classifier was used. Generally, the proposed method demonstrates better accuracy and f-values compared to other methods. 

 Compared to naive Bayes classifiers, with Bayesian networks, it is difficult to search for and find optimal structures while expressing complex structures. An exhaustive search causes a combinatorial explosion even with a small number of nodes; thus, in this study, the structure search is conducted heuristically. In addition, accuracy and f-values tend to be unstable because an exhaustive search may result in local solutions. Therefore, in the proposed method, in consideration of the probability that data are associated with multiple learners, predictions are made using learners with probability higher than a given threshold. This prediction method enables robust prediction of the structure of Bayesian networks. Furthermore, if only the optimal Bayesian network structure can be found, the proposed method demonstrates higher accuracy than the comparison method. 

The proposed method can improve accuracy by retraining according to the prediction result after learning the Bayesian networks and multiple learners. In future, it will be necessary to verify how retraining affects accuracy.
In addition, the proposed method requires more computation time than using a single deep learner because deep learners must be trained relative to the number of clusters. The computation time of multiple learners is limited by the learners of clusters with large amounts of data. However, it is possible to reduce computation time significantly by exploiting parallel calculation. Thus, in future, we plan to implement parallel computing to improve the efficiency of the proposed method.

In addition, in future we intend to develop a method that uses a Bayesian network to integrate outputs from multiple learners. In the current proposed method, there are two approaches to combine a Bayesian network with multiple learners. One approach is to apply a Bayesian network prior to inputting the data to multiple deep learners, and the other is to apply a Bayesian network when integrating the outputs from multiple learners. In this paper, we have described both approaches and detailed the method based on the former approach. However, by combining these approaches, we expect that it will be possible to design a method to integrate the outputs from multiple learners using deep learners, clustering, and a Bayesian network. Thus, in future we would like to develop a method to select learners based on the data and make predictions using the selected learners.

\section*{Conflict of interest}
The authors declare that they have no conflict of interest.

\bibliographystyle{unsrt}


\end{document}